\documentclass[conference]{IEEEtran}
\IEEEoverridecommandlockouts
\usepackage{cite}
\usepackage{amsmath,amssymb,amsfonts}
\usepackage{algorithmic}
\usepackage{graphicx}
\usepackage{textcomp}
\usepackage{xcolor}
\usepackage{cuted}
\usepackage{float}
\usepackage{times}
\usepackage{epsfig}
\usepackage{placeins}
\usepackage{verbatim}
\usepackage{multirow}
\usepackage{multicol}
\def\BibTeX{{\rm B\kern-.05em{\sc i\kern-.025em b}\kern-.08em
    T\kern-.1667em\lower.7ex\hbox{E}\kern-.125emX}}
\begin{document}

\title{Learning task-specific features for 3D pointcloud graph creation}

\author{\IEEEauthorblockN{Elias Abad Rocamora}
\IEEEauthorblockA{\textit{Image Processing Group} \\
\textit{Universitat Politècnica de Catalunya}\\
Barcelona, Spain \\
abad.elias00@gmail.com}
\and
\IEEEauthorblockN{Javier Ruiz-Hidalgo}
\IEEEauthorblockA{\textit{Image Processing Group} \\
\textit{Universitat Politècnica de Catalunya}\\
Barcelona, Spain \\
j.ruiz@upc.edu}}

\maketitle

\begin{abstract}
Processing 3D pointclouds with Deep Learning methods is not an easy task. A common choice is to do so with Graph Neural Networks, but this framework involves the creation of edges between points, which are explicitly not related between them. Historically, naive and handcrafted methods like k Nearest Neighbors (k-NN) or query ball point over xyz features have been proposed, focusing more attention on improving the network than improving the graph. In this work, we propose a more principled way of creating a graph from a 3D pointcloud. Our method is based on performing k-NN over a transformation of the input 3D pointcloud. This transformation is done by an Multi-Later Perceptron (MLP) with learnable parameters that is optimized through backpropagation jointly with the rest of the network. We also introduce a regularization method based on stress minimization, which allows to control how distant is the learnt graph from our baseline: k-NN over xyz space. This framework is tested on ModelNet40, where graphs generated by our network outperformed the baseline by 0.3 points in overall accuracy.
\end{abstract}

\begin{IEEEkeywords}
3D pointclouds, Graph Neural Networks
\end{IEEEkeywords}

\section{Introduction}
\label{sec:Intro}

A 3D pointcloud is a set of data points in a space. In their most basic form, every data point contains just spatial coordinates xyz, but can contain other features as RGB color or surface normals. This set of data points may represent any object like a car, a plane, a chair or a guitar \cite{ModelNet40}. But can also be a set of points sampled from the surface of objects in a real world scene \cite{S3DIS}, \cite{ScanObjectNN} with coordinated cameras or LiDAR sensors. Similarly as one does with standard images, one might be interested in classifying, segmenting or detecting objects in a 3D pointcloud, but this is quite more challenging as we will explain.

In \cite{DL_3DPC_Review}, authors define 3 main challenges of processing 3D pointclouds with Deep Learning (DL) methods:
\begin{itemize}
  \item Irregularity: Points might not be sampled uniformly around the object or scene they represent.
  \item Unstructuredness: Each point is sampled individually and the distance between points is not fixed, in contrast, regular 2D images are captured in a grid and distance between adjacent pixels is always the same.
  \item Unorderdness: The 3D pointcloud is a set of points, meaning that the order in which each point is stored, doesn't change the object or scene it is representing.
\end{itemize}

In order to tackle these, different approaches have been proposed.

\begin{figure}[t]
\begin{center}
   \includegraphics[width=1\linewidth]{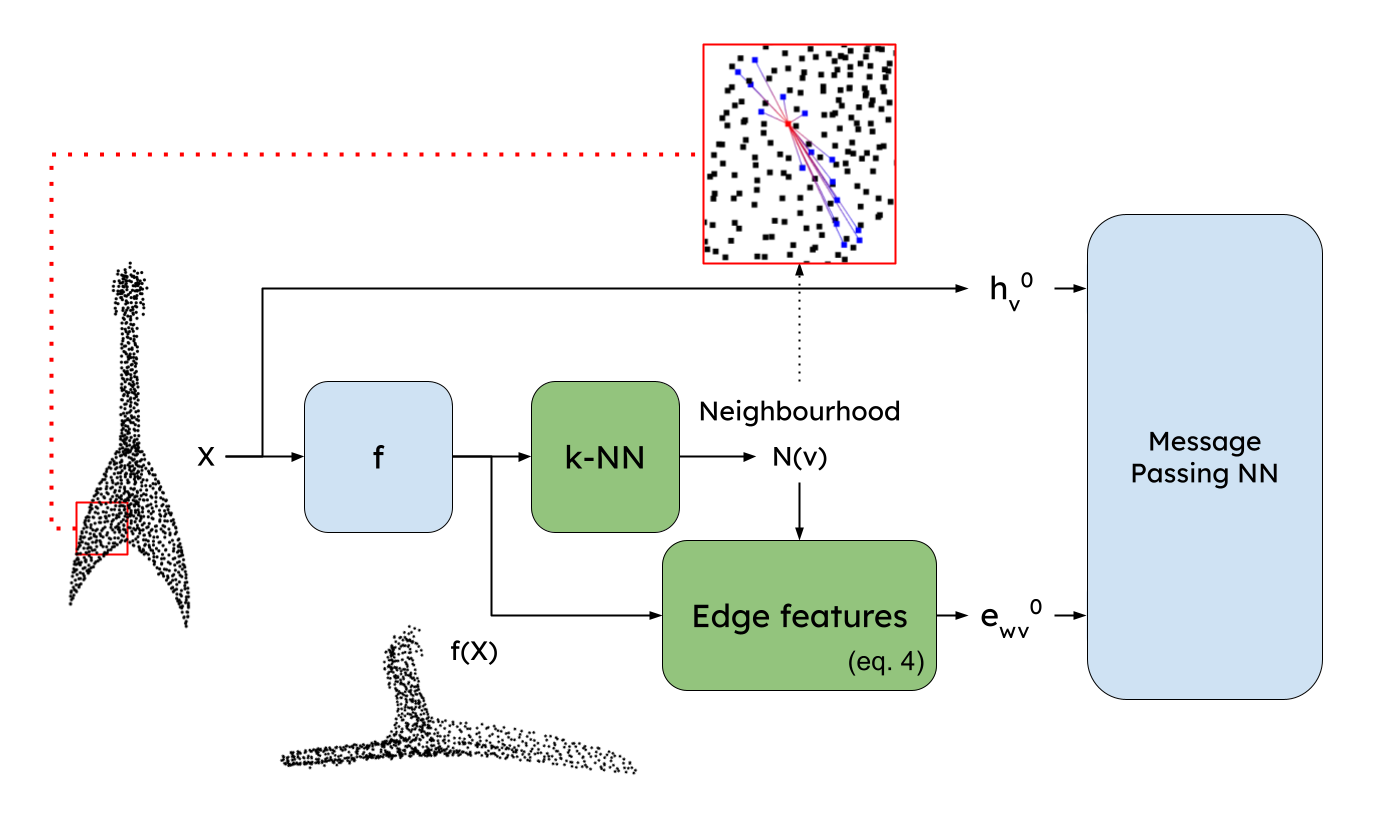}
\end{center}
   \caption{Graph creation pipeline: Each point is mapped from initial features to another space via $f$, where k-NN is performed to create the neighborhood. A combination of both nodes' transformed features is used to produce edge features, which along original node features are passed to a Message Passing Neural Network.}
\label{fig:SimPip}
\end{figure}

\textit{Multiview based}: These methods take advantage of the very advanced area of Convolutional Neural Networks (CNN) on 2D images. 3D pointclouds are projected onto 2D grids from one or more points of view and then processed with a CNN architecture \cite{MultiviewCNN}, \cite{end-to-end_multiview}, \cite{VNN2019}. 

\textit{Voxel based}: Similarly to pixels in 2D images, voxels are 3D cubes that form a 3D grid. Voxel based methods, represent 3D pointclouds with voxels in a offline processing step and then apply convolutions with 3D kernels to process the voxelized 3D pointcloud \cite{PointGrid}, \cite{3DConv}.

\textit{Directly processing 3D pointclouds}: Multiview and Voxel based methods rely on projections onto 2D and 3D grids to solve Irregularity, Unstructuredness and Unorderedness. But by projecting onto grids, exact point-level information is lost. PointNet \cite{PointNet} was the first architecture able to directly process unstructured 3D point clouds. This was done by applying two symmetric functions: a Multi-Later Perceptron (MLP) with learnable parameters and a maxpooling operation to obtain a global feature vector of the 3D pointcloud. This architecture, however, does not take advantage of local information around every point. The local dependence that is captured with 2D or 3D convolutions is desired for unstructured 3D pointclouds. Subsequent work combines PointNet ideas with locality aware operations to improve state-of-the-art.

A way of providing a structure to the pointcloud and make use of local correlation is with the use of Graph Neural Networks (GNN). Methods processing 3D pointclouds with GNN, define a graph where each point is a node and add directed edges by joining each point with its nearby ones. But as the relationship between points is not explicit, some mechanism needs to be applied to define the graph edges. Some authors propose joining each point with its k Nearest Neighbors (k-NN) \cite{3DGNN}, \cite{PointGCN} or with the points at a distance within a radius $\rho$ (query ball point) \cite{GACNET}, both in the xyz space, regardless of other features the pointcloud might have. More recent approaches \cite{DeepGCN}, \cite{DGCNN} use features learnt at every layer to dynamically change the neighbourhood with k-NN, but still use xyz features in the first layer. Dynamically changing the neighborhood with learnt node features has been proven to improve performance on 3D pointcloud tasks, but little effort has been put on exploring new methods different than k-NN or query ball point, or which features other than xyz are better for creating the edges of the graph.

In this work, we focus on learning the right features to create the graph with k-NN. Whe propose a method that generalizes k-NN by mapping every point with an MLP taking all its features and performing k-NN on that new space. Whe also introduce a regularization term that controls how far our graph is from k-NN over xyz features, and test everything over the popular 3D pointcloud classification benchmark ModelNet40 \cite{ModelNet40}.

\section{Related work}
\label{sec:RelWork}

There are numerous works that rely on Graph Neural Networks to process unstructured 3D point clouds in different manners \cite{kd_trees}, \cite{3DGNN},\cite{DGCNN}, \cite{Point2Node}, \cite{GACNET}, \cite{DeepGCN}. All these architectures construct graphs by defining a node for each point in the point cloud and use some rule to define edges between them. Reference \cite{3DGNN} creates the graph via k-NN over xyz features and applies a GNN involving an MLP to transform input node fratures, and an LSTM \cite{LSTM} which taks previous node features as hidden state and sum of transformed neighbor features as an input to produce new node features. Reference \cite{kd_trees} uses kd-tree graph and builds a feed-forward kd-network, where the leaves are the input of the network and the root is the output where the graph representation is encoded. Reference \cite{DGCNN} creates the edges by joining every node with k-NN over xyz, then applies a GNN updating every node features as a maxpooling operation of incoming edges' features. Learnt features are used to built the graph again, at every layer, based on k-NN in the new feature space. At every layer, \cite{GACNET} creates the edges by joining every node with k randomly chosen points that are at distance $ \text{d} \leq \rho_{l}$ in xyz, where $\rho_{l}$ increases at every layer $l$, node features are updated by using an attention mechanism similar to \cite{GAT} over adjacent nodes and furthest point sampling is used to reduce the dimensionality. Reference \cite{DeepGCN} uses a GCN aggregating neighbors' features with maxpooling a operation, and an MLP taking aggregation of neighbors' and previous node features to produce new ones. This is done at d-dilated k-NN graphs, in the first layer, xyz features are used and at consequent layers, in the same way as \cite{DGCNN}, learnt node features are used.

\section{Method}
\label{sec:Meth}

Motivated by how \cite{DGCNN} and \cite{DeepGCN} use features at every layer to dynamically change the neighborhood, but all methods still use xyz features for the first layer graph, we designed our method to learn the best possible features to create the graph in the first layer. For this work, we decided to keep the same graph at every layer as we will focus on comparing it with regular k-NN over xyz features.

\subsection{Message Passing Neural Network}
\label{subsec:MPNN}

Our network relies on the Message Passing Neural Network (MPNN) framework \cite{MessagePassing}. This framework generalizes many Graph Neural Network methods with two phases: message passing phase, which runs for T timesteps and updates the node features, and a readout phase, which computes a feature vector for the whole graph.

\subsubsection{Message passing}
\label{subsubsec:MP}
We base our work on the MPNN case defined in \cite{Kearnes2016}. In this work, edge features are introduced into the MPNN. At each message passing (MP) block, the node features of a node $v$ are updated with an MLP that takes as an input the concatenation ($. \mathbin\Vert .$) of the previous hidden state $h^{t-1}_{v}$ and the message $m^{t}_{v}$:

\begin{equation}\label{eq:node_update}
h^{t}_{v} = H^{t}(h^{t-1}_{v} \mathbin\Vert m^{t}_{v})
\end{equation}

Where $H^{t}$ is an MLP and the message $m^{t}_{v}$ is the sum of the features of its incoming edges:

\begin{equation}\label{eq:message_update}
m^{t}_{v}=\sum_{w \in N(v)}e^{t}_{wv}
\end{equation}

See \ref{subsec:GraphCreation} for our definition of $N(v)$. Edge features $e^{t}_{wv}$ are updated with an MLP $E^{t}$ that takes as an input the concatenation of $e^{t-1}_{wv}$, $h^{t-1}_{w}$ and $h^{t-1}_{v}$:

\begin{equation}\label{eq:edge_update}
e^{t}_{wv} = E^{t}(e^{t-1}_{wv} \mathbin\Vert h^{t-1}_{w} \mathbin\Vert h^{t-1}_{v})
\end{equation}

Initial node hidden states $h^{0}_{v}$ are the pointcloud spatial features (xyz) and possibly other features like RGB color. We follow \cite{DGCNN} and define starting edge features $e^{0}_{wv}$ as:

\begin{equation}\label{eq:initial_edge_feats}
e^{0}_{wv} = (f(h^{0}_{v}) \mathbin\Vert f(h^{0}_{w}) - f(h^{0}_{v}))
\end{equation}

$f$ can be any mapping.

\begin{figure*}[t]
\begin{center}
   \includegraphics[width=1\linewidth]{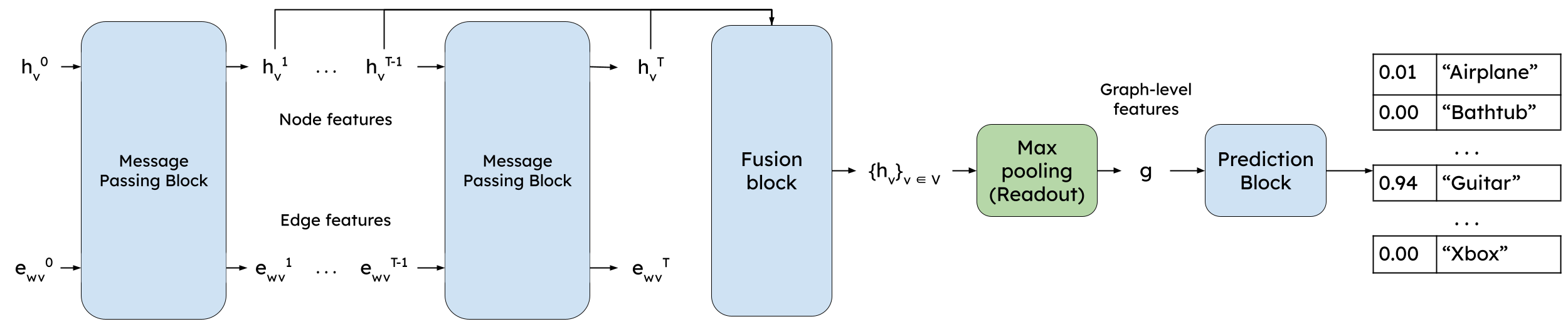}
\end{center}
   \caption{Message Passing Neural Network architecture for the graph classification task. After T message passing blocks, node features from each timestamp are aggregated with a fusion block, obtaining a final feature vector for each node. Then, max pooling is performed across all nodes, to obtain a feature vector of the whole graph. Finally, a prediction block with a final softmax layer is applied to output the class probability distribution.}
\label{fig:ComPipe}
\end{figure*}

\subsubsection{Readout phase}
\label{subsubsec:Read}

Lastly, in the readout phase, we aggregate all the node features and obtain a global feature vector $g$ of the graph, see Fig. \ref{fig:ComPipe}. In order to do so, we need an order-invariant function like sum or max, in this work we apply max pooling after a fusion block with a skip connection and fully connected layer, to aggregate features from each Message Passing block and obtain a final feature vector $h_{v}$ for each node:

\begin{equation}\label{eq:readout}
g = \max_{v \in V} h_{v} = \max_{v \in V}G(h^{1}_{v} \mathbin\Vert h^{2}_{v} \mathbin\Vert ... \mathbin\Vert h^{T}_{v})
\end{equation}

Where $G$ is an MLP and $V$ is the set of nodes in the graph. Finally, a set of fully connected layers can be applied to map $g$ to the desired number of classes. If our task is to provide an output for each one of the nodes in the graph, the readout phase can be avoided and apply a shared weight MLP to each node features.

\subsection{Latent features for k-NN edge creation}
\label{subsec:GraphCreation}

As mentioned in related work \ref{sec:RelWork}, many proposed architectures create directed graph edges by joining every node with its k-NN on input space. We propose learning a mapping $f:\mathbb R^{d_{\textit{in}}} \to\mathbb R^{d_{\textit{graph}}}$ to another space, where we will perform k-NN on every node to create the edges of our graph, see Fig. \ref{fig:SimPip}. Our mapping $f$ is defined by an MLP with learnable parameters $F$. Finally, the set of neighbors $N(v)$ of a node $v$ is assigned as the k-NN in euclidean distance from the features $f(h^{0}_{v})$.

\begin{comment}
\begin{equation}\label{eq:k-nn}
N(v) = \text{k-NN}_{f(h^{0}_{w}), w \in V}(f(h^{0}_{v}))
\end{equation}
\end{comment}

\subsection{Stress Loss}
\label{subsec:Stress}

Stress is a well known metric in the statistics community. It was defined in \cite{Kruskal1964} and it is the criterion to optimize when performing Multidimensional Scaling (MDS). We use Stress as a regularization of our mapping function, by jointly minimizing Stress, we restrict our MLP from mapping the input points to a complex space where the distances between points are very different. In this sense, Stress captures how well the transformed points represent the actual points. Let $d_{ij} = {\lVert x_{j} - x_{i} \rVert}_2$ and $\hat{d}_{ij} = {\lVert f(x_{j}) - f(x_{i}) \rVert}_2$ for $x_i, x_j \in \mathcal{X} = [x_1, x_2, ..., x_n] \in \mathbb{R}^3$ being the input pointcloud in xyz and $f:\mathbb{R}^3 \to \mathbb{R}^d$ the mapping represented by our MLP, Stress is defined as follows:

\begin{equation}\label{eq:stress}
    S(d,\hat{d}) = \sqrt{\frac{\sum_{i<j}(d_{ij} - \hat{d}_{ij})^2}{\sum_{i<j}d_{ij}^2}}
\end{equation}

Then, to avoid computing the outer squared root, we can optimize squared stress, and do it jointly with the criterion of our task:

\begin{equation}\label{eq:loss}
    \min_{\theta} L = L_{task}(\theta) + \gamma S(d,\hat{d}(\theta))^2
\end{equation}

Where $\theta$ are the parameters of our model, $L_{task}$ the appropriate loss function for our graph learning task (Classification, regression, etc.) and $\gamma$ is a scalar fixed weight to control the importance of stress in the optimization of $L$.

\section{Experiments}
\label{sec:Exp}

We test our work on a widely used 3D point cloud classification benchmark: ModelNet40 \cite{ModelNet40}. For all of our experiments, we use an architecture with $T=4$ Message passing stages. As defined in section \ref{subsubsec:MP}, each Message Passing (MP) block is composed of two MLPs: $H^{t}$ and $E^{t}$, see eq. \ref{eq:node_update} and \ref{eq:edge_update} respectively. The input size of $E^{t}$, depends on the number of output features of $F^{t}$, we denote this number as $d_{\textit{graph}}$, see section \ref{subsec:GraphCreation}. Both $H^{t}$ and $E^{t}$ depend on the number of input node features $d_{\textit{in}}$. Finally, we use an MLP $P$ with $d_{\textit{classes}}$ output dimensions (40 for ModelNet40) after the readout phase to predict the pointcloud class. For the graph creation, we use $k=16$ for selecting the k-NN. The whole architecture is trained through back-propagation for 100 epochs, with batch size equal to 16, a learning rate of $10^{-4}$ which is decayed in half every 20 epochs, ADAM optimizer \cite{kingma2017adam} and Cross-entropy Loss for $L_{task}$, see \ref{eq:loss}. During the training process, weights from the epoch with the highest validation accuracy are kept and used for providing results. Every result provided is the average of 3 runs with different random initialization seeds.

In order to test the proposed architecture, we compare with a baseline model using $f(x_{i}) = xyz$, which is equivalent to using the Identity function for Modelnet40, as no other node features, such as color ones, are provided in this dataset. Therefore, $d_{\textit{graph}} = d_{\textit{in}} = 3$ . 
%This baseline also allows us to be compared with other methods that create the neighborhood via k-NN over the spatial features.

\subsection{Effect of $d_{\textit{graph}}$}

We investigate which is the best number of features to represent each point before performing k-NN and entering the first MP block. In order to do so, we build $F$ as a 2-layer MLP with a ReLU activation between them. The output dimensions of the first layer is fixed to 16 and 6 different values of $d_{\textit{graph}}$ (1,2,3,6,9 and 12) are tested in terms of average class accuracy and overall accuracy over the ModelNet40 dataset. For this experiment $\gamma$ was set to 0 for every model.

\begin{table}[t]
\caption{Classification results on ModelNet40 for different values of $d_{\textit{graph}}$}
\begin{center}
\begin{tabular}{|l|c|c|c|c|c|}
\hline
\multicolumn{1}{|p{1cm}|}{\centering Model} & \multicolumn{1}{|p{1cm}|}{\centering $d_{\textit{graph}}$} & \multicolumn{1}{|p{1cm}|}{\centering Stress} &  \multicolumn{1}{|p{1cm}|}{\centering overall \\ Acc.}\\
\hline\hline
baseline & 3 & 0 & 91.4\\
\hline
mlp-3-16-1 & 1 & 0.496  & 89.9\\
mlp-3-16-2 & 2 & 0.482  & 90.3\\
mlp-3-16-3 & 3 & 0.674 & 91.2\\
mlp-3-16-6 & 6 & 1.12  & 91.2\\
mlp-3-16-9 & 9 & 1.89  & 91.3 \\
mlp-3-16-12 & 12 & 2.61  & \bf{91.5}\\
\hline
\end{tabular}
\end{center}
\label{tab:d_graph}
\end{table}

\begin{comment}
\begin{table}[t]
\caption{Classification results on ModelNet40 for different values of $d_{\textit{graph}}$}
\begin{center}
\begin{tabular}{|l|c|c|c|}
\hline
\multicolumn{1}{|p{1cm}|}{\centering Model} & \multicolumn{1}{|p{1cm}|}{\centering $d_{\textit{graph}}$} & \multicolumn{1}{|p{1cm}|}{\centering overall \\ Acc.}\\
\hline\hline
baseline & 3 & 91.4\\
\hline
mlp-3-16-1 & 1 & 89.9\\
mlp-3-16-2 & 2 & 90.3\\
mlp-3-16-3 & 3 & 91.2\\
mlp-3-16-6 & 6 & 91.2\\
mlp-3-16-9 & 9 & 91.3 \\
mlp-3-16-12 & 12 & \bf{91.5}\\
\hline
\end{tabular}
\end{center}
\label{tab:d_graph}
\end{table}
\end{comment}

For output dimensions smaller than the input, the performance is clearly affected, with 1.5 and 1.1 points less in overall accuracy for $d_{\textit{graph}} = 1$ and $d_{\textit{graph}} = 2$ respectively. This result is natural as F needs to find the configuration that provides best accuracy, and at the same time reduce the dimensionality of the data. In this case, the model cannot be able to learn the identity function as it is limited in the number of dimensions, the best it can do to be close to the baseline is to find a non-linear 1D/2D projection where distances in 3D are approximated.

For $d_{\textit{graph}} \geq 3$, F is able to learn an identity mapping and therefore should be able to attain at least the same performance as our baseline. This is however not observed in our experiments, only for $d_{\textit{graph}} = 12$ we are able to outperform the baseline in overall accuracy by 0.1 points \ref{tab:d_graph}. This makes us think that the training of these models is not converging to the best solution possible. In this sense, the use of $\gamma = 0$ might be a too unrestricted scenario for F to converge.

\subsection{Effect of $\gamma$}

With the $\gamma$ parameter, we can explicitly control how much importance has Stress in the optimization of $F$ parameters, we argue that if k-NN over spatial features is one of the most widely used graph creation method and has proven to provide excellent results, solutions that resemble well distances in the input space, should also provide good performance. Additionally, the fact that on the mlp-3-16-3 setup, $F$ did not converge to the identity, but neither provided better or similar results to the baseline, made us think that $F$ was not converging to the best possible solution. In order to analyze the effect of $\gamma$, we trained the best performing setup from our first experiment, mlp-3-16-12, and the endomorphism setup mlp-3-16-3 with $\gamma$ values ranging from 0.0001 to 10 in x10 increments. 

\begin{table}[t]
\caption{Stress, percentage of edges shared with baseline's graph and Classification results on ModelNet40 for different values of $\gamma$}
\begin{center}
\begin{tabular}{|c|c|c|c|c|c|c|}
\hline
\multicolumn{1}{|p{0.9cm}|}{\centering Model} & \multicolumn{1}{p{1cm}|}{\centering $\gamma$} & \multicolumn{1}{p{1cm}|}{\centering Stress} & \multicolumn{1}{p{1.5cm}|}{\centering \% shared \\ edges} & \multicolumn{1}{p{1cm}|}{\centering overall \\ Acc.}\\
\hline\hline
baseline &  &  &  & 91.4\\
\hline
%\multirow{7}{*}{\multicolumn{1}{p{0.9cm}}{\centering mlp \\ 3-16-12}} & 0 & 2.61 & 83.0 & 91.5 \\
& 0 & 2.61 & 83.0 & 91.5 \\
& 0.0001 & 1.48 & 83.2 & 91.2\\
mlp & 0.001 & 0.438 & 84.7 & 91.5 \\
3-16-12 & 0.01 & 0.168 & 85.0 & 91.2 \\
& 0.1 & 0.0396 & 93.1 & 91.3 \\
& 1 & 0.0115 & 96.4 & 91.3 \\
& 10 & 0.00305 & 98.4 & 91.5 \\
\hline
%\multirow{7}{*}{\multicolumn{1}{p{0.9cm}}{\centering mlp \\ 3-16-3}}& 0 & 0.674 & 54.2 & 91.2\\
& 0.0001 & 0.578 & 58.9 & 91.2 \\
& 0.001 & 0.331 & 57.8 & 91.2\\
mlp & 0.01 & 0.179 & 67.6 & 91.2\\
3-16-3 & 0.1 & 0.0233 & 93.3 & 91.3\\
& 1 & 0.0071 & 97.2 & 91.6\\
& 10 & 0.00223 & 99.4 & \bf{91.7} \\
\hline
\end{tabular}
\end{center}
\label{tab:d_gamma}
\end{table}

\begin{figure*}
\begin{center}
\includegraphics[width=0.9\linewidth]{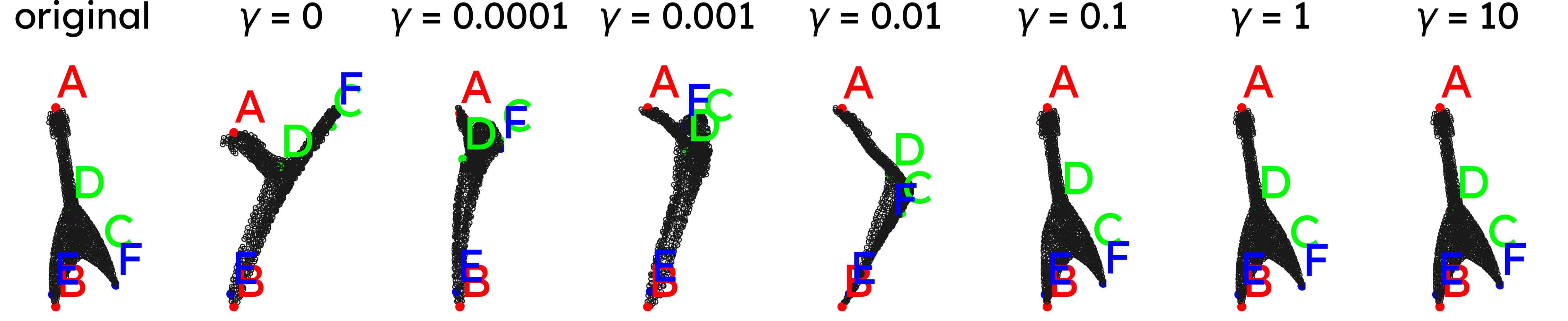}
\includegraphics[width=0.9\linewidth]{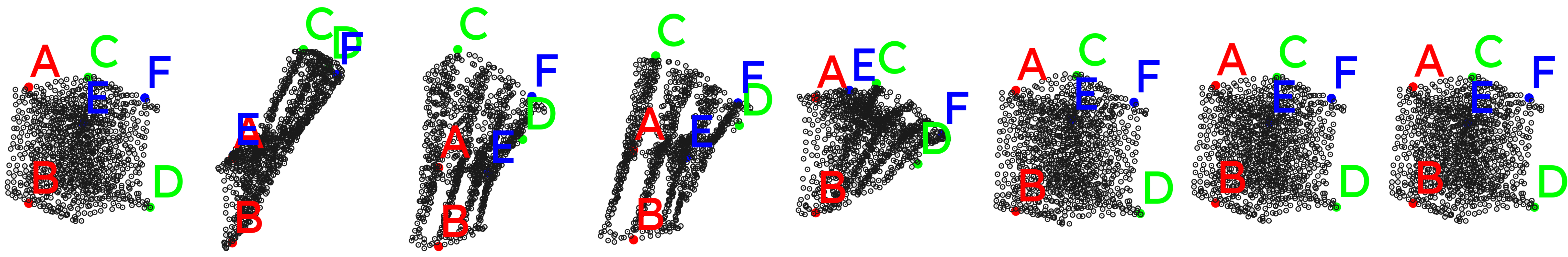}
\end{center}
   \caption{2D projection of ModelNet40 test pointclouds number 29 and 512 after being transformed with $F$ in mlp-3-16-3 setup trained with different $\gamma$ values. "A" and "B","C" and "D" and "E" and "F" points correspond to argmax and argmin of x, y and z axis respectively in the original pointcloud. Every 3D pointcloud is projected into the hyperplane defined by "A", "B" and "F" in each space.}
\label{fig:projections}
\end{figure*}

The higher the values of $\gamma$, the more $F$ preserves distances and therefore, the more the generated graph is similar to the baseline's. This is observed in terms of Stress, which our experiments show to decrease with bigger $\gamma$s and in terms of percentage of edges shared with baseline graph, which increases with $\gamma$ until reaching a value of 98.4 and 99.4 for $d_{\textit{graph}} = 12$ and $d_{\textit{graph}} = 3$ respectively at $\gamma = 1$, meaning an almost perfect match with baseline graph, see table \ref{tab:d_gamma}. This is confirmed when looking at the plots of some of the transformed pointclouds. If one takes a look at Fig. \ref{fig:projections}, one can observe that on rightmost columns, corresponding to highest values of gamma, the pointclouds are very similar to the original one, while at the letmost columns, one is not able to distinguish visually some of the shapes.

With the highest values of $\gamma$, we are able to converge to approximately the same graph as the baseline, but surprisingly obtain better accuracy results, 91.5 and 91.7 for $d_{\textit{graph}} = 12$ and $d_{\textit{graph}} = 3$ respectively at $\gamma = 10$. This is in fact because we are using the same graph, but different edge features. the only thing that separates our mapping $F$ from the identity is rotation. By minimizing Stress, we are only enforcing that the point-to-point distances are preserved. This means that for high $\gamma$ values, $F$ is just applying a rotation to the input pointcloud, and therefore our network takes as an input the original rotation of the pointcloud via the initial node features $h_v^{0}$, and the rotation applied by $F$ via the edge features $e_{vw}^0$, see equation \ref{eq:initial_edge_feats}. As CGNs on their own are not rotation invariant operators, the network "looking at" the pointcloud in two different orientations, might explain the improvement in accuracy.

\section{Conclusions and future work.}
\label{sec:Concl}

In this work, we have proposed a new framework for creating graphs from 3D pointclouds that is integrated into the optimization of the GNN. We also propose a regularization method in order to control how far is the generated graph from the k-NN over xyz solution by Stress minimization. The proposed architecture is not able to learn graphs that provide significantly better results than our baseline by its own, only when regularizing with our proposed method, we are able to surpass the baseline by 0.3 overall accuracy points. But the generated graph is the same as the baseline's and features used to create the graph are just a rotation of the input pointcloud.

We believe that regardless this work not improving significantly our baseline, we have layed a foundation on where future work can be built. It is still not clear if any other neighborhood selection method different from k-NN or query ball point performs better on 3D pointclouds, or if other features different from xyz might work better with k-NN or query ball point. In this work, when not optimizing Stress ($\gamma = 0$), learnt features are non easily interpretable, further work is needed on analyzing which properties do they have and why does our method converge to them. Another limitation of our method is that the size of each node's neighborhood is fixed to k. Investigating more general methods that generate varying size neighborhoods, at the same time as learning the best node-level features to do so, could be a promising research line.

\bibliographystyle{ieee_fullname}
\bibliography{egbib}

\begin{thebibliography}{10}\itemsep=-1pt

\bibitem{S3DIS}
Iro Armeni, Ozan Sener, Amir~R. Zamir, Helen Jiang, Ioannis Brilakis, Martin
  Fischer, and Silvio Savarese.
\newblock 3d semantic parsing of large-scale indoor spaces.
\newblock In {\em Proceedings of the IEEE International Conference on Computer
  Vision and Pattern Recognition}, 2016.

\bibitem{DL_3DPC_Review}
Saifullahi~Aminu Bello, Shangshu Yu, and Cheng Wang.
\newblock Review: deep learning on 3d point clouds.
\newblock {\em CoRR}, abs/2001.06280, 2020.

\bibitem{MessagePassing}
Justin Gilmer, Samuel~S. Schoenholz, Patrick~F. Riley, Oriol Vinyals, and
  George~E. Dahl.
\newblock Neural message passing for quantum chemistry.
\newblock {\em CoRR}, abs/1704.01212, 2017.

\bibitem{Point2Node}
Wenkai Han, Chenglu Wen, Cheng Wang, Xin Li, and Qing Li.
\newblock Point2node: Correlation learning of dynamic-node for point cloud
  feature modeling.
\newblock {\em CoRR}, abs/1912.10775, 2019.

\bibitem{VNN2019}
Xinwei He, Tengteng Huang, Song Bai, and Xiang Bai.
\newblock View n-gram network for 3d object retrieval.
\newblock {\em CoRR}, abs/1908.01958, 2019.

\bibitem{LSTM}
Sepp Hochreiter and Jürgen Schmidhuber.
\newblock Long short-term memory.
\newblock {\em Neural computation}, 9:1735--80, 12 1997.

\bibitem{Kearnes2016}
Steven Kearnes, Kevin McCloskey, Marc Berndl, Vijay Pande, and Patrick Riley.
\newblock Molecular graph convolutions: moving beyond fingerprints.
\newblock {\em Journal of Computer-Aided Molecular Design}, 30(8):595–608,
  Aug 2016.

\bibitem{kingma2017adam}
Diederik~P. Kingma and Jimmy Ba.
\newblock Adam: A method for stochastic optimization, 2017.

\bibitem{kd_trees}
Roman Klokov and Victor~S. Lempitsky.
\newblock Escape from cells: Deep kd-networks for the recognition of 3d point
  cloud models.
\newblock {\em CoRR}, abs/1704.01222, 2017.

\bibitem{Kruskal1964}
J.~B. Kruskal.
\newblock Multidimensional scaling by optimizing goodness of fit to a nonmetric
  hypothesis.
\newblock {\em Psychometrika}, 29(1):1--27, Mar 1964.

\bibitem{PointGrid}
Truc Le and Ye Duan.
\newblock Pointgrid: A deep network for 3d shape understanding.
\newblock In {\em 2018 IEEE/CVF Conference on Computer Vision and Pattern
  Recognition}, pages 9204--9214, 2018.

\bibitem{DeepGCN}
Guohao Li, Matthias M{\"{u}}ller, Ali~K. Thabet, and Bernard Ghanem.
\newblock Can gcns go as deep as cnns?
\newblock {\em CoRR}, abs/1904.03751, 2019.

\bibitem{end-to-end_multiview}
Lei Li, Siyu Zhu, Hongbo Fu, Ping Tan, and Chiew{-}Lan Tai.
\newblock End-to-end learning local multi-view descriptors for 3d point clouds.
\newblock {\em CoRR}, abs/2003.05855, 2020.

\bibitem{3DConv}
Daniel Maturana and Sebastian Scherer.
\newblock 3d convolutional neural networks for landing zone detection from
  lidar.
\newblock In {\em 2015 IEEE International Conference on Robotics and Automation
  (ICRA)}, pages 3471--3478, 2015.

\bibitem{PointNet}
Charles~Ruizhongtai Qi, Hao Su, Kaichun Mo, and Leonidas~J. Guibas.
\newblock Pointnet: Deep learning on point sets for 3d classification and
  segmentation.
\newblock {\em CoRR}, abs/1612.00593, 2016.

\bibitem{3DGNN}
Xiaojuan Qi, Renjie Liao, Jiaya Jia, Sanja Fidler, and Raquel Urtasun.
\newblock 3d graph neural networks for rgbd semantic segmentation.
\newblock In {\em 2017 IEEE International Conference on Computer Vision
  (ICCV)}, pages 5209--5218, 2017.

\bibitem{MultiviewCNN}
Hang Su, Subhransu Maji, Evangelos Kalogerakis, and Erik~G. Learned{-}Miller.
\newblock Multi-view convolutional neural networks for 3d shape recognition.
\newblock {\em CoRR}, abs/1505.00880, 2015.

\bibitem{ScanObjectNN}
Mikaela~Angelina Uy, Quang{-}Hieu Pham, Binh{-}Son Hua, Duc~Thanh Nguyen, and
  Sai{-}Kit Yeung.
\newblock Revisiting point cloud classification: {A} new benchmark dataset and
  classification model on real-world data.
\newblock {\em CoRR}, abs/1908.04616, 2019.

\bibitem{GAT}
Petar Veličković, Guillem Cucurull, Arantxa Casanova, Adriana Romero, Pietro
  Liò, and Yoshua Bengio.
\newblock Graph attention networks, 2018.

\bibitem{GACNET}
Lei Wang, Yuchun Huang, Yaolin Hou, Shenman Zhang, and Jie Shan.
\newblock Graph attention convolution for point cloud semantic segmentation.
\newblock In {\em 2019 IEEE/CVF Conference on Computer Vision and Pattern
  Recognition (CVPR)}, pages 10288--10297, 2019.

\bibitem{DGCNN}
Yue Wang, Yongbin Sun, Ziwei Liu, Sanjay~E. Sarma, Michael~M. Bronstein, and
  Justin~M. Solomon.
\newblock Dynamic graph {CNN} for learning on point clouds.
\newblock {\em CoRR}, abs/1801.07829, 2018.

\bibitem{ModelNet40}
Zhirong Wu, Shuran Song, Aditya Khosla, Xiaoou Tang, and Jianxiong Xiao.
\newblock 3d shapenets for 2.5d object recognition and next-best-view
  prediction.
\newblock {\em CoRR}, abs/1406.5670, 2014.

\bibitem{PointGCN}
Yingxue Zhang and Michael Rabbat.
\newblock Ieee international conference on acoustics, speech and signal
  processing (icassp).
\newblock In {\em 2017 IEEE International Conference on Computer Vision
  (ICCV)}, pages 6279--6283, 04 2018.

\end{thebibliography}

\end{document}